\documentclass{article}

\usepackage[final,nonatbib]{neurips_2019}

\usepackage[utf8]{inputenc} % allow utf-8 input
\usepackage[T1]{fontenc}    % use 8-bit T1 fonts
\usepackage{hyperref}       % hyperlinks
\usepackage{url}            % simple URL typesetting
\usepackage{booktabs}       % professional-quality tables
\usepackage{amsfonts}       % blackboard math symbols
\usepackage{nicefrac}       % compact symbols for 1/2, etc.
\usepackage{microtype}      % microtypography

\usepackage{graphicx}
\usepackage{subfig}
\usepackage{wrapfig}
\usepackage{titlesec}

\title{It's easy to fool yourself: Case studies on identifying bias and confounding in bio-medical datasets}

\titlespacing\section{0pt}{10pt plus 4pt minus 2pt}{0pt plus 2pt minus 2pt}
\titlespacing\subsection{0pt}{12pt plus 4pt minus 2pt}{0pt plus 2pt minus 2pt}

\author{%
 Subhashini Venugopalan$^{*1}$, Arunachalam Narayanaswamy$^{*1}$, Samuel Yang\thanks{equal contribution}$\ \ ^{1}$, \\ 
 {\textbf{Anton Geraschenko}$^{1}$}, 
 {\textbf{Scott Lipnick}$^{2}$}, 
 {\textbf{Nina Makhortova}$^{2}$}, 
 {\textbf{James Hawrot}$^{3}$}, 
 {\textbf{Christine Marques}$^{3}$}, \\
 {\textbf{Joao Pereira}$^{3}$},
 {\textbf{Michael Brenner}$^{1,2}$}, 
 {\textbf{Lee Rubin}$^{2}$}, 
 {\textbf{Brian Wainger}$^{3}$}, 
 {\textbf{Marc Berndl}$^{1}$} \\
 {{$^{1}$Google} {$^{2}$Harvard} {$^{3}$Massachusetts General Hospital
 }}
}

\begin{document}

\maketitle

\begin{abstract}
Confounding variables are a well known source of nuisance in biomedical studies. They present an even greater challenge when we combine them with black-box machine learning techniques that operate on raw data. This work presents two case studies. In one, we discovered biases arising from systematic errors in the data generation process. In the other, we found a spurious source of signal unrelated to the prediction task at hand. 
In both cases, our prediction models performed well but under careful examination hidden confounders and biases were revealed. These are cautionary tales on the limits of using machine learning techniques on raw data from scientific experiments.
\end{abstract}
\section{Introduction}
Deep learning provides powerful tools to unravel hidden signal in data. The field has had tremendous success applying it to a number of problems in the medical domain from detecting cancers \cite{esteva2017dermatologist,liu2017detecting}, diabetic retinopathy~\cite{gulshan2016development}, to predicting cardio vascular risk factors~\cite{poplin2018prediction}. 
But along with the power comes the peril of hunting for the source of signal in our data. In experimental sciences, confounders pose well known pitfalls. While using hand-engineered features extracted from raw observations, these have been less of a concern and models built on them have fewer parameters to learn spurious signal. However, deep neural nets built on raw high dimensional data such as images have greatly amplified the ability of these models to exploit confounding variables. For example, any variation in image acquisition settings such as camera type, background noise levels, illumination can be quickly exploited by these models while carefully hand-engineered features might be immune to a certain degree. These factors are usually not causally connected to the prediction task at hand.

In this work, we show some pitfalls in using deep neural nets on microscopic images of cells. We have used deep neural net models to find novel biomarkers for identifying cell types. We show how some biases could be identified by careful experimental design, visualization of model outputs and low dimensional projections of embeddings. We also employed a few model interpretability techniques to drill down to the source of our signal in our target classification model. Our work covers two case studies that try to identify biomarkers for: (i) Spinal Muscular Atrophy (SMA) from human fibroblast cells and (ii) Amyotrophic Lateral Sclerosis (ALS) from induced pluripotent stem cell (iPSC) derived motor neurons.
\section{Data generation} 
\begin{figure}%
    \centering
    \subfloat[Data acquisition]{{\includegraphics[width=0.4\textwidth]{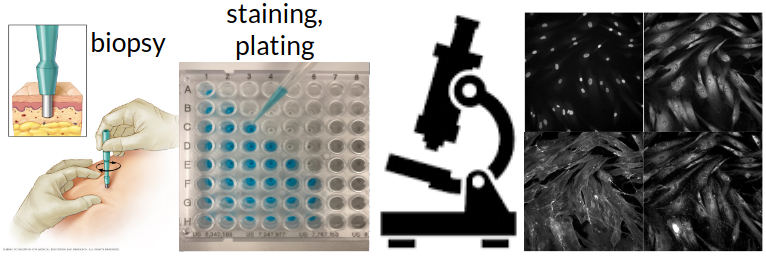} }}%
    \quad
    \subfloat[Preprocessing]{{\includegraphics[width=0.5\textwidth]{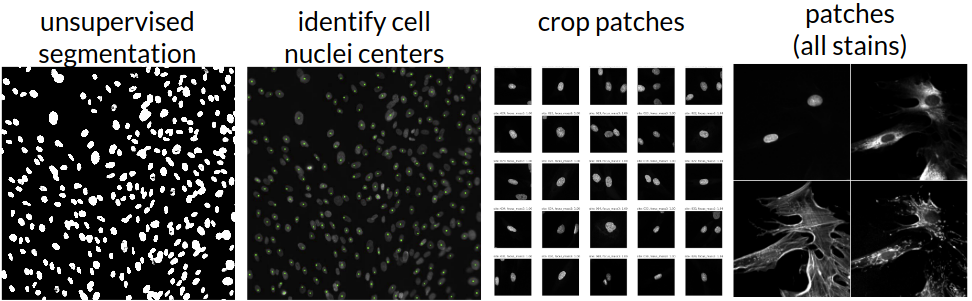} }}%
    \caption{\small{Data generation starts with acquiring a piece of tissue or cells from a person. The cells are cultured, plated, stained, and then imaged using a microscope. The images are processed by segmenting cells (unsupervised), identifying cell centers, and cropping centered patches.}}%
    \label{fig:data_acq}%
\end{figure}

Data generation (Figure \ref{fig:data_acq}) starts with acquiring a piece of tissue or cells from a person (cell line). Cells are cultured, plated, then fixed, and stained.  
They are imaged through a high throughput microscope at 5 frequency bands (which are analogous to channels) at various sites (physical locations) in a well. The same experimental setup is replicated on several plates at multiple time-points a few weeks apart. Each time-point replicate is termed ``batch" (or experimental batch). A typical experimental batch could contain 12 plates, each plate containing 96 wells, each imaged at multiple locations/sites. Each image has 5 channels. % and each containing 10-100s of cells per site. 
In our datasets, each well on a plate contains cells for one experimental condition, i.e. fibroblasts or iPSC cells from a cell line from a single individual. These cell lines were strategically distributed across the plates to mitigate known plate covariate nuisance effects.

\textbf{Pre-processing}
From a microscope image acquired at a given site, there may be 10 to 100 individual cells in the image. We follow the unsupervised segmentation approach from \cite{mandopaper}
to detect cell nuclei and obtain fixed-sized crops centered around the nuclei 
yielding the cell ``patch" image. Image analysis can be done with either the entire site image, or the cell patch images.

\vspace{-3pt}
\subsection{Datasets}
For this work we used two types of datasets: SMA and ALS. The first  one is from human fibroblast cells to detect spinal muscular atrophy from \cite{yang2019applying}, split into two versions: a pilot dataset (SMA pilot dataset) and the main dataset (SMA Main dataset). 
For ALS, we obtained images of motor neurons differentiated from iPSC (Cell preparation similar to the one in \cite{wainger2014intrinsic}) 
from healthy / isogenic pairs TDP43 introduced. 
Cells were allowed to mature after plating and imaged after 3-5 days. The cells were fixed, stained and imaged in both the cases through a process similar to the one described in \cite{bray2016cell}.

\textbf{SMA Pilot Dataset}
Six primary fibroblast cell lines were distributed and cultured in wells on 12 96-well plates in a single experimental batch. 
 This was primarily used to study and correct biases in the acquisition process.

\textbf{SMA Main Dataset}
27 primary fibroblast cell lines from 12 subjects affected by SMA, and 15 otherwise healthy demographically matched (age, race, sex) subjects were used in an experiment with two batch replicates conducted weeks apart
, yielding about 2 million images in total.

\textbf{ALS dataset}
For this dataset, we had 5 independently differentiated batches that were imaged at 9 sites per well. Each batched contained 1-2 plates containing 96 wells per plate containing two types of cells: TDPwt and TDPmut representing a isogenic pair of motor neurons with and without ALS. We split the dataset into 5 folds leaving one batch out in each fold. Hence each fold contained 4 batches in train and 1 in validation. We did not use a separate test set for these experiments.
\vspace{-4pt}
\section{Approach}
\vspace{-2pt}
\textbf{Quality analysis}
We first evaluate the focus quality of the images using a pre-trained CNN model described in \cite{yang2018assessing}. The model is trained on biological images that are artificially blurred to different extents. It uses a ranked probability score to determine overall focus quality of the image and predicts a rating with a numeric score ranging from 0 to 1 (1 indicating in-focus). Figure~\ref{fig:res_focus_qlty} visualizes the scores in the spatial layout of the acquired site images in wells on the 96-well plates.

\textbf{Detecting nuisance using unsupervised embeddings}
For a first pass at identifying biases in the data, we use unsupervised \textbf{clustering}, t-SNE, to identify nuisance variables visually (Figure~\ref{fig:res_tsne}). We apply t-SNE on image embeddings of cell patches obtained from an inceptionv4~\cite{szegedy2017inception} model pre-trained on ImageNet~\cite{deng2009imagenet}.
We label the points in the resulting space by various known covariate factors, such as plate and location within a plate. %
Additionally, we also quantitatively assess and predict these nuisance covariate factors using \textbf{logistic regression} (Appendix Figure~\ref{fig:res_nuis_pred})

\textbf{Supervised prediction of disease condition}
To assess whether the disease state of the subjects 
can be inferred from the microscopy images of the cells, a k-fold cross validation scheme was used, whereby each fold contained a test set of a pair (one from a healthy individual, the other from a disease individual) of cell lines unseen in the training set. We use logistic classifiers and CNNs (using modified inceptionv4 with predicted heads) %
to predict the ``disease condition" of the cell line. 
The logistic classifiers were applied on unsupervised embeddings, and also a set of 63 hand-engineered features computed from image statistics such as foreground area and foreground mean intensity, etc.

\textbf{Model Interpretation}
We used visual explanation tools to understand which regions in the image our deep learning model is most influenced by when predicting healthy or disease. %
We present the saliency maps obtained by GradCAM~\cite{selvaraju2017grad} in Figure~\ref{fig:res_gcam_prow}.
We also use Partial Dependency Plots (PDP)~\cite{breiman2001random} on hand-engineered features to learn associations between target features and target responses.

\vspace{-10pt}
\section{Results}
\vspace{-5pt}
\subsection{Detecting bias from quality analysis visualization}
\begin{figure}[h]

  \centering
  \includegraphics[width=\textwidth]{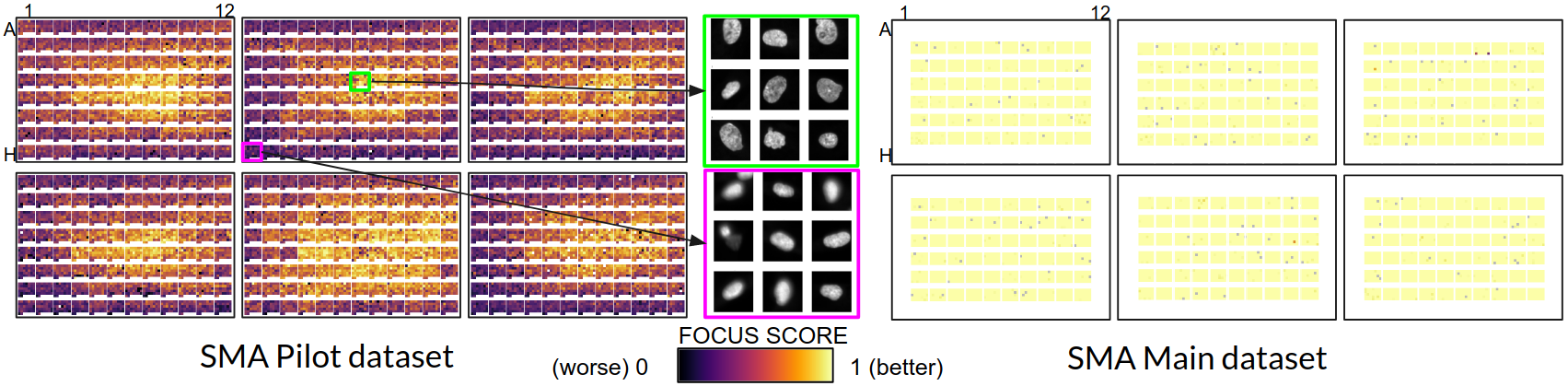}
  \caption{\small{Visualization of the results from applying the focus quality analysis on 6 96-well plates from the SMA Pilot dataset (left) and the SMA Main dataset (right). Each pixel corresponds to a site within a well, and the color represents the focus quality, yellow being in-focus and purple depicting out-of-focus. 9 sample cell images from 2 wells are presented in the center. %
  }}
  \label{fig:res_focus_qlty}
  
\end{figure}

As Figure~\ref{fig:res_focus_qlty} shows, the detected focus quality in our preliminary SMA pilot dataset reveals a clear spatial trend, whereby the images of the cells located towards the center of each 96-well plate during the experiment had better focus quality. To address this we acquired a z-stack of confocal images instead of a single widefield image which was used to create the SMA Main dataset. This helped generate images with much higher focus. %

\vspace{-2pt}
\subsection{Detecting bias based on unsupervised clustering}
\begin{figure}[h]%
    \centering
    \subfloat[SMA Main dataset]{{\includegraphics[width=0.4\textwidth]{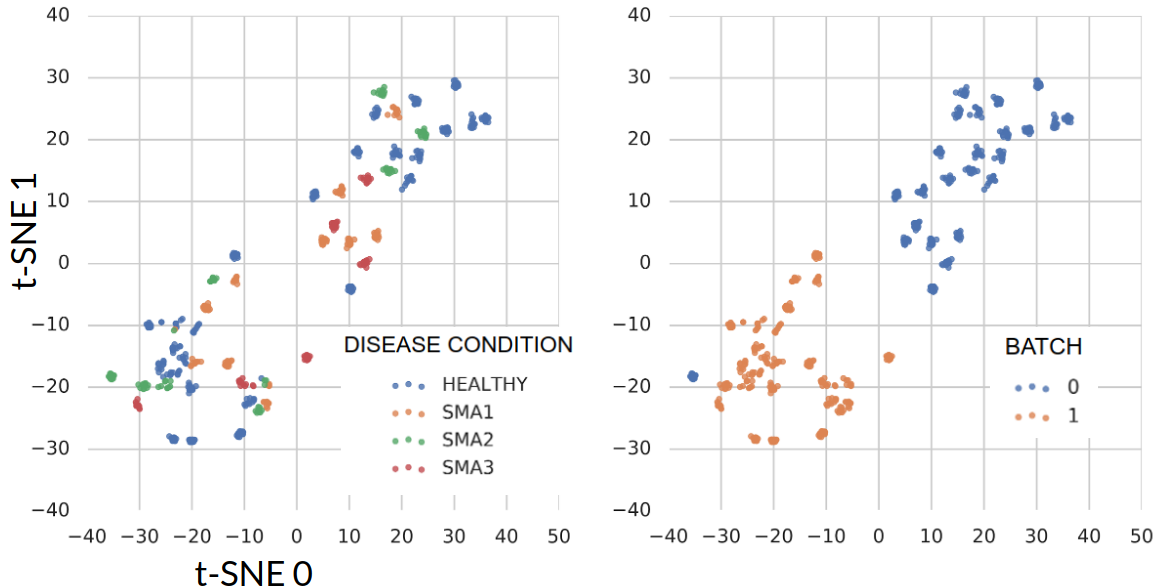} }}%
    \quad
    \subfloat[ALS  dataset]{{\includegraphics[width=0.5\textwidth]{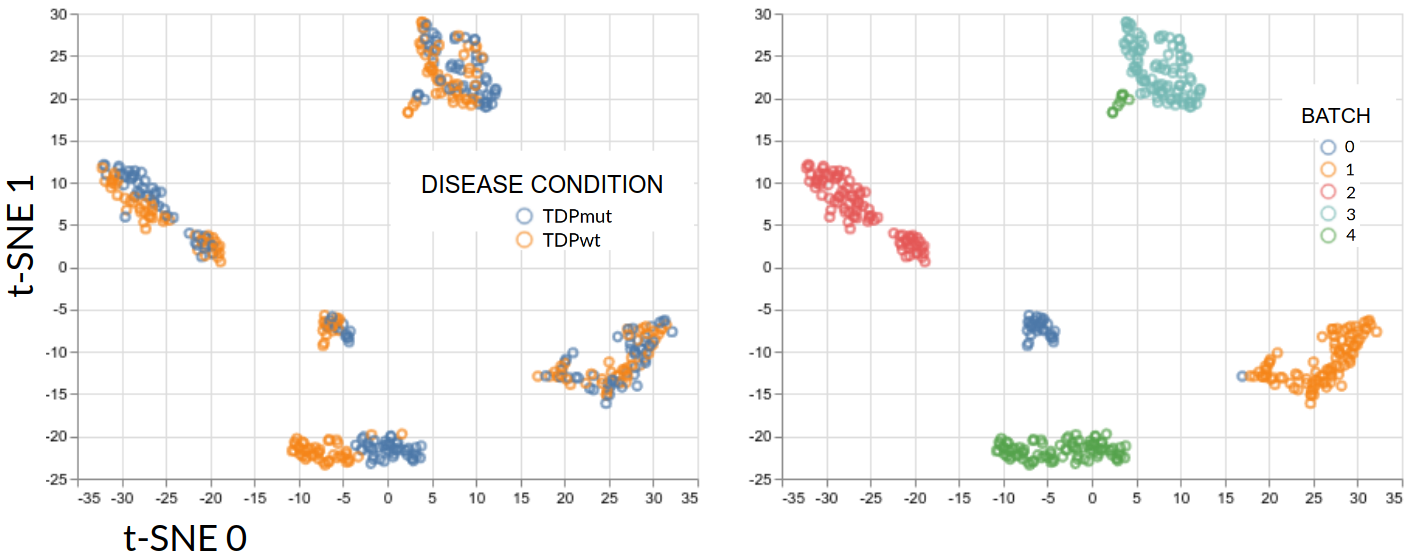} }}%
    \caption{\small{Clustering on unsupervised CNN embeddings using t-SNE on two datasets. In each image, the data points are colored based on either disease condition or the batch.}}%
    \label{fig:res_tsne}%
\end{figure}

\vspace{-2pt}
Results from clustering of the unsupervised embeddings in Figure~\ref{fig:res_tsne} show that in both the SMA Main dataset and the ALS dataset, there is a strong bias for images to cluster based on the experimental batch. This was also confirmed by quantitative assessment (appendix Figure~\ref{fig:res_nuis_pred}).

\vspace{-2pt}
\subsection{Examination of supervised prediction models reveal confounders}
\begin{figure}[h]
  \centering
  \subfloat[Supervised models]{{\includegraphics[width=0.35\textwidth,height=3.3cm]{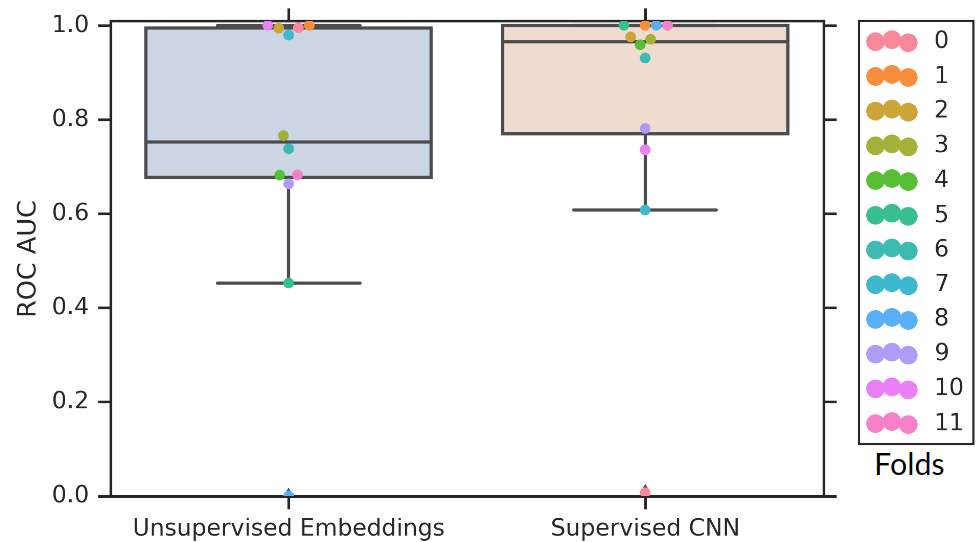} }}%
    \quad
    \subfloat[Supervised CNN: Examining the folds]{{\includegraphics[width=0.6\textwidth]{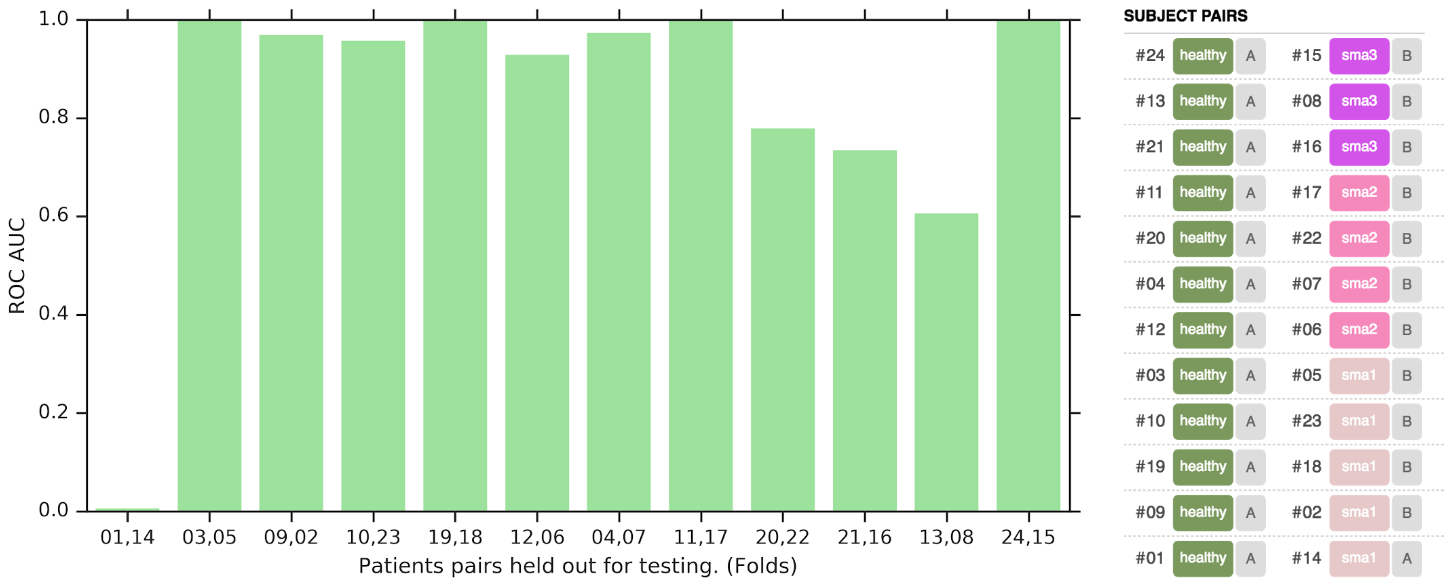}}}%
  \caption{\small{Plots comparing performance of supervised classifiers to predict healthy and disease lines on the SMA Main dataset. (a) Comparison of unsupervised embedding (logistic classifier) and supervised CNN. Each point represents ROC AUC performance on one cross-validation fold. (b) Performance of the fully supervised CNN on the SMA Main dataset on each of the 12 folds (held-out patient pairs). (right) The disease condition (healthy, sma1-3) and the lab source (A or B, grey square) where the patient's cells were acquired.}
  }
  \label{fig:res_sma_w_pred}
\end{figure}
\vspace{-4pt}
\paragraph{Predict healthy/disease}
We sought to assess whether the disease state of individuals from which the cells originated could be inferred from the cell images, for cells from an unseen person. The results on the SMA Main dataset, as shown in Figure~\ref{fig:res_sma_w_pred}, indicate that on an unseen pair of cell lines from an unseen healthy and disease individual, the model is able to predict with better-than-chance accuracy in all but one of the cross validation folds.

While promising, one significant covariate is the source lab of cell lines; the only cross validation fold with worse-than-random generalization is also the one in which both the healthy and disease cell line were obtained from the same, rather than different, lab sources. As seen in Figure~\ref{fig:res_sma_w_pred}, either the cell lines in this fold were outliers, or the selectivity of the model may depend on a combination of cell line source and disease state.
\begin{wrapfigure}[13]{r}{0.4\textwidth}
    \centering
     \includegraphics[width=0.4\textwidth]{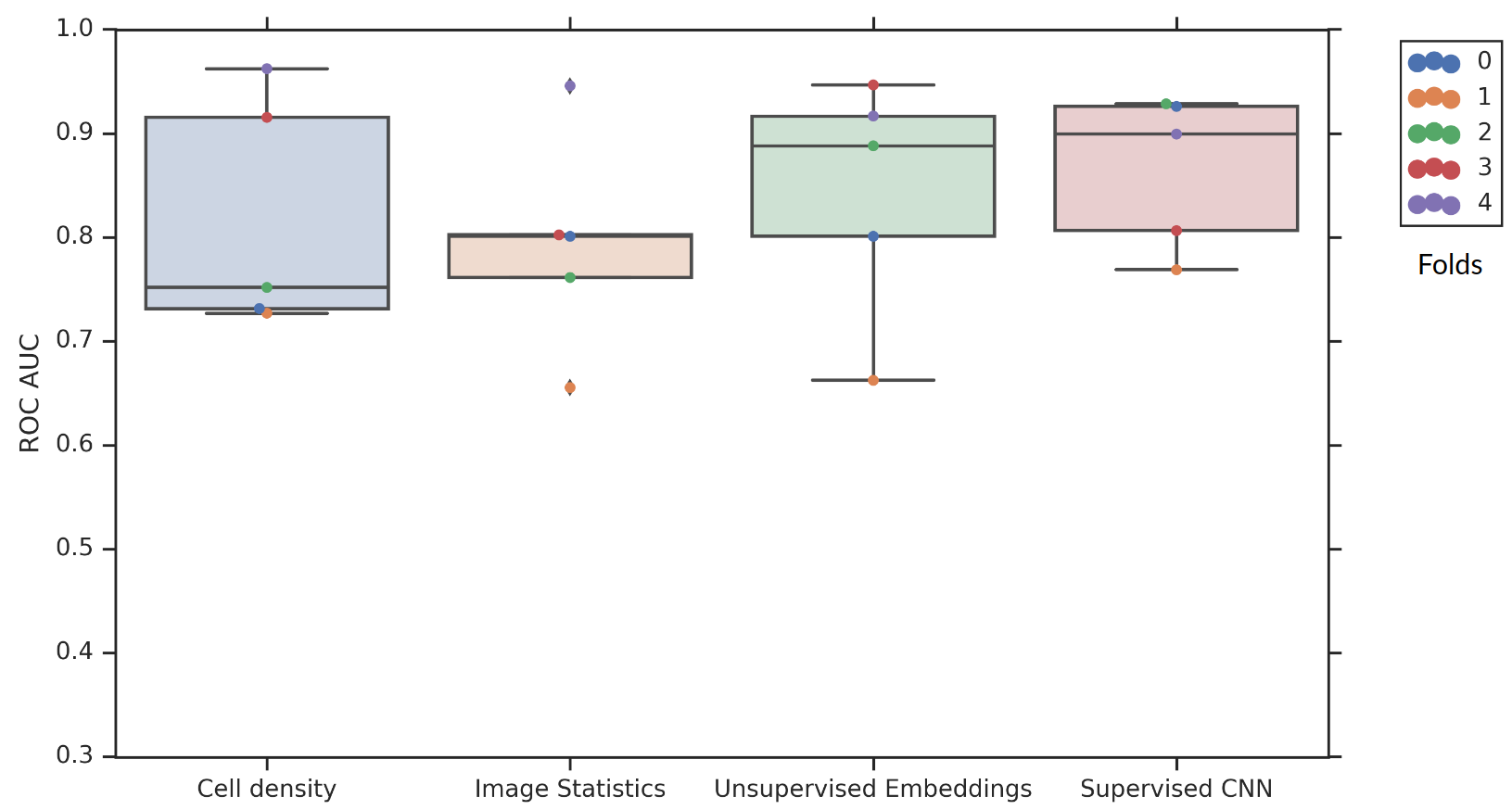}%
    \caption{\small{[ALS Dataset] Comparison of logistic regression model trained on  cell density, image statistics features, unsupervised embeddings, and the fully supervised CNN models. Each point represents ROC AUC performance on one cross-validation fold.}}%
    \label{fig:res_sma_qnt}%
\end{wrapfigure}

\textbf{ALS Dataset} On the ALS dataset (Figure~\ref{fig:res_sma_qnt}) while the supervised CNN model performs the best; the model based on 63 hand-engineered features from the image statistics, and a model that uses just the cell density alone (i.e. number of cells in the site image) performs comparably. Partial dependency plots (PDP)~\cite{breiman2001random} applied on the  hand-engineered features pointed to cell density as the confounder. %

\subsection{Model interpretation points to confounding}
\begin{figure}[h]
  \centering
  \includegraphics[width=\textwidth]{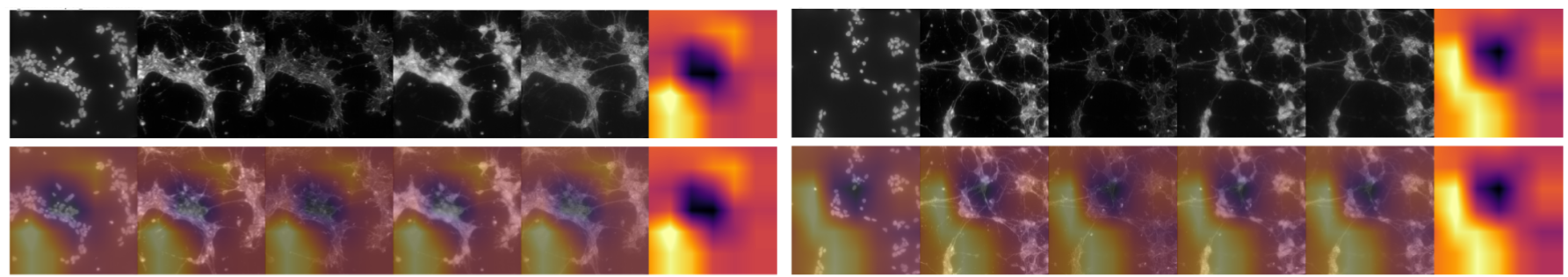}
  \caption{\small{Saliency maps from GradCAM overlaid on the different image channels/stains for 2 sets of images from the ALS dataset. Yellow regions represent where the model is ``looking" to correctly predict TDPwt cell lines.}}
  \label{fig:res_gcam_prow}
\end{figure}

Saliency maps obtained by applying GradCAM on the ALS dataset are presented in Figure~\ref{fig:res_gcam_prow}. The saliency maps indicate that the model is looking at empty regions when correctly predicting lines to be of type TDPwt. This further indicates density of cells as a confounding variable that the models exploit.
\section{Conclusion}
In our work, we observed our models exploiting differences in the cell line source, experimental batch, plate, relative location of wells in a plate, image acquisition settings, cell density etc. to identify the batch / plate / well containing the input and thereby gaining insight into the possible target label. We then used these deep neural nets to understand the focus quality differences among various wells. We have used supervised classifiers to identify the extent to which the models can memorize nuisance factors and also employed model interpretability techniques to provide further evidence of the source of (spurious) signal in our trained models. Our work emphasizes the need to be cognizant of these pitfalls and urge the community to carefully examine the source of signal, especially in case of a novel discovery.
\section*{Acknowledgements}
The authors would like to thank Minjie Fan, Zan Armstrong, Thorsten M. Schlaeger, Liyong Deng, Wendy K. Chung, Liadan O'Callaghan, Dosh Whye, Jon Hazard, Brian Patrick Williams, D. Michael Ando, and Philip Nelson for their help with earlier versions of this work.

\small{
\bibliographystyle{abbrv}
\bibliography{references}
}
\newpage
\section*{Appendix}
\begin{figure}[hbt]
  \centering
  \includegraphics[width=\textwidth]{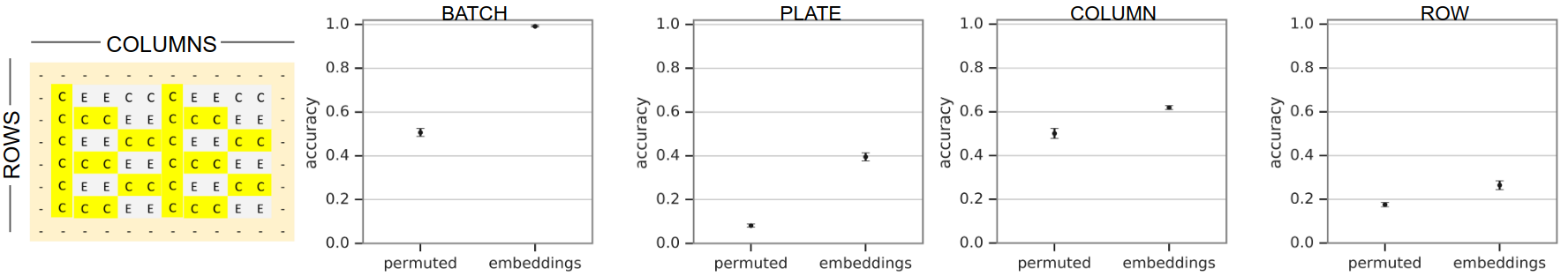}
  \caption{\small{Comparing prediction of nuisance factors - batch, plate, row, column - on the SMA Main dataset. (leftmost) Depicts a plate with experimental (E) wells, and control (C) wells highlighted in yellow. (right plots) Logistic regression models on unsupervised CNN embeddings of cells from the "control" wells with that of the embedding feature columns permuted (as the baseline).}}
  \label{fig:res_nuis_pred}
\end{figure}

\textbf{Identifying bias by predicting nuisance covariates}
We used the unsupervised embeddings of cell images from the "control" wells to predict nuisance factors such as batch, plate, and well position using logistic regression. We also compare this to predicting the nuisance factors by permuting the embeddings feature columns~\cite{breiman2001random}, and present the results for the SMA Main dataset in Figure~\ref{fig:res_nuis_pred}. As seen in the figure, there is a higher than chance accuracy of predicting these nuisance variables indicating the underlying bias in the data.

\end{document}